\documentclass[sigconf,nonacm]{acmart}
\AtBeginDocument{%
  }

\usepackage{algorithm}
\usepackage{algorithm}
\usepackage{algorithmic}
\usepackage{makecell}
\usepackage{multirow}
\usepackage{balance}
\setcopyright{none}
\renewcommand\footnotetextcopyrightpermission[1]{}

\begin{document}

\title{TUGB: A Two-stage Uncertainty-aware Generative Auto-Bidding Framework}
\thanks{This paper has been accepted by CIKM 2026.}

\author{Luzhan Yuan}
\email{yuanluzhan.1@jd.com}
\authornotemark[1]
\affiliation{%
  \institution{JD.com}
  \city{Beijing}
  \country{China}
}

\author{Chao Mi}
\authornote{Both authors contributed equally to this research.}
\email{michao3@jd.com}
\affiliation{%
  \institution{JD.com}
  \city{Beijing}
  \country{China}
}

\author{Xiahao Shi}
\email{shixiahao@jd.com}
\affiliation{%
  \institution{JD.com}
  \city{Beijing}
  \country{China}
}

\author{Yi Mao}
\email{maoyi3@jd.com}
\affiliation{%
  \institution{JD.com}
  \city{Beijing}
  \country{China}
}

\author{Shengsheng Niu}
\email{niushengsheng@jd.com}
\affiliation{%
  \institution{JD.com}
  \city{Beijing}
  \country{China}
} 

\author{Xin Zhu}
\email{zhuxin3@jd.com}
\affiliation{%
  \institution{JD.com}
  \city{Beijing}
  \country{China}
}

\author{Jie He}
\email{hejie67@jd.com}
\affiliation{%
  \institution{JD.com}
  \city{Beijing}
  \country{China}
}

\author{Zhangang Lin}
\email{linzhangang@jd.com}
\affiliation{%
  \institution{JD.com}
  \city{Beijing}
  \country{China}
}

\author{Ching Law}
\email{lawching@jd.com}
\affiliation{%
  \institution{JD.com}
  \city{Beijing}
  \country{China}
}
\renewcommand{\shortauthors}{Luzhan Yuan et al.}

\begin{abstract} 
Generative modeling, particularly Decision Transformers, shows promise in auto-bidding strategies. However, it struggles with limited exploration in static offline datasets, vulnerability to dynamic environmental shifts, and a reliance on heuristic return-to-go (RTG) specifications that may yield suboptimal performance on unseen traffic. To address these challenges, we propose the Two-stage Uncertainty-aware Generative Auto-Bidding (TUGB) framework. TUGB integrates generative sequence modeling with Reinforcement Learning (RL). It employs a two-stage pipeline where a pre-trained Generative Bidding Model is fine-tuned via an ensemble-based Uncertainty-aware Environment Model. This approach quantifies epistemic uncertainty to penalize high-risk trajectories and evolves the policy from RTG-conditioning to direct bidding optimization. Extensive experiments demonstrate that TUGB significantly outperforms SOTA baselines. Notably, full-scale deployment on JD Search Ad achieved a 2.32\% increase in cost and 3.57\% in target cost, validating its industrial efficacy, while its successful adaptation to subsequent mechanism upgrades and JD Food Delivery Ad further validates TUGB's superior generalization capability.
\end{abstract}




\maketitle

\section{Introduction}
In modern online advertising systems, such as search engines, social media platforms, and e-commerce platforms, auto-bidding has emerged as the primary tool for campaign management~\cite{onlineAD, RTB1, RTB2, AutoBiddingSurvey}. It allows advertisers to go beyond manually managing bids for enormous ad impressions. Instead, advertisers simply define high-level business objectives (e.g., maximize conversions, target Return on Investment (tROI) or target a cost-per-acquisition (tCPA)) and set constraints like budget, cost per click, etc. 
For each ad request, the auto-bidding agent dynamically computes a bid by combining these objectives with real-time contextual signals and user characteristics. This bid is then submitted to the real-time auction, determining whether the ad is displayed and the price paid based on market competition. Therefore, auto-bidding serves as a critical bridge in the ad system: it translates an advertiser's macroscopic business target into impression level, per-auction bidding actions, aiming to maximize the advertiser's outcome within a complex, varying bidding environment.

Recently, generative modeling has emerged as a powerful paradigm for auto-bidding, with the Decision Transformer (DT) and its variants serving as representative frameworks~\cite{AIGB, DT, GAS, GAVE, likaiyuanEBaReT}. These methods cast auto-bidding as a sequence generation task, where decision-making is conditioned on Return-to-Go (RTG) and state trajectories. By specifying a target RTG, the model generates bidding actions based on learned historical patterns. However, the effectiveness of DT-based methods is often compromised by the non-stationary nature of real-world auction environments. First, current DT-based methods are trained on offline datasets collected via existing online policies, which inherently limits their exploration capabilities and prevents the model from reliably extrapolating to policies that outperform the data source~\cite{xiyaopolicygap,offlineRLsurvey,GapinofflineRL}. 
Second, dynamic shifts in auction environments introduce epistemic uncertainty, as the model struggles to generalize to unseen market states. This uncertainty creates a critical risk: the agent may overestimate the value of these out-of-distribution (OOD) states, leading to aggressive bidding decisions that undermine campaign performance~\cite{OODdanger}. Last but not least, determining an optimal RTG is non-trivial in complex auctions; an improper RTG setting—especially on unseen traffic where the maximum potential return is unknown and can lead to suboptimal or even detrimental actions~\cite{RTGbad, RTGbad2}. While recent approaches like GAS and GAVE attempt to enhance exploration~\cite{GAS, GAVE}, they still rely heavily on precise RTG settings and lack the uncertainty-awareness to mitigate overestimation in policy optimization.

To address these challenges, we propose a Two-stage Uncertainty-aware Generative Auto-Bidding Framework (TUGB). TUGB comprises two key components: a Generative Bidding Model (GBM) designed to model bidding distributions and drive decision-making, and an Uncertainty-aware Environment Model (UEM) that facilitates the optimization of the GBM while quantifying environmental uncertainty. Through an Uncertainty-aware Bidding Optimization fine-tuning stage, the framework empowers the agent to actively explore beyond static offline datasets while leveraging the UEM to quantify epistemic uncertainty and penalize high-risk trajectories. This design effectively evolves the bidding policy from heuristic Return-to-Go (RTG) conditioning to risk-aware optimal auto-bidding decision-maker,  enabling dynamic adaptation without oracle dependencies. 
TUGB has been fully deployed on the JD Search Ad, achieving a 2.32\% increase in cost and a 3.57\% increase in target cost, validating its industrial efficacy. Furthermore, TUGB has been adapted to the subsequent ad mechanism upgrade and JD Food Delivery Ad. These results demonstrate that TUGB not only significantly boosts cost and target cost but also exhibits superior ability to adapt different bidding target and data sparse scenario.

The main contributions of this paper are summarized as follows:
\begin{itemize} 
    \item \textbf{TUGB Framework}:
    We propose TUGB, a two-stage framework bridging generative modeling with uncertainty-aware reinforcement learning for auto-bidding. TUGB addresses three critical challenges in generative auto-bidding: the high financial risk of exploration, epistemic uncertainty arising from non-stationary market dynamics, and the reliance on heuristic RTG targets. By leveraging an Uncertainty-aware Environment Model (UEM) to explicitly quantify uncertainty and penalize high-risk trajectories, TUGB facilitates safe exploration beyond offline data, enabling the generative bidding model to transform from a passive imitator into a risk-aware optimal auto-bidding decision-maker.
    
    \item \textbf{Industrial Deployment and Generalization}:
    TUGB has been fully deployed on JD.com's search advertising system, achieving 2.32\% in cost and 3.57\% in target cost while maintaining strict constraint satisfaction. Beyond static deployment, the framework demonstrates robust cross-scenario transferability: (i) it rapidly adapts to system mechanism upgrades via reward function changing without architectural modifications; (ii) it generalizes to JD Food Delivery through few-shot fine-tuning, leveraging the pre-trained policy prior to overcome distributional shifts. 
\end{itemize}

\section{Preliminaries}
In this section, we provide the foundational background necessary to understand our proposed framework. We formalize the auto-bidding task as a constrained optimization problem. We then briefly introduce generative decision-making methods, with a specific emphasis on their practical applications in auto-bidding scenario and their RL-integrated variants.

\subsection{Auto-Bidding Problem Formulation}

In automated bidding systems, advertisers typically establish long-term performance objectives—such as a target Cost Per Acquisition (tCPA) or target Return on Investment (ROI) subject to a specific budget constraint over a defined period. To achieve these goals, the bidding agent continuously interacts with the ad auction mechanism, making sequential bidding decisions for each arriving impression.

For example, we consider an advertiser aiming to deliver an ad campaign on JD.com (a leading e-commerce platform). The advertiser specifies a target cost per acquisition ($\text{tCPA}$) and allocates a total budget $B$ for the campaign. In an evaluation period, the campaign involves a sequence of $N$ requests. For each request $i$, let $x_i \in \{0, 1\}$ denote the binary allocation variable, where $x_i=1$ indicates winning the impression at cost $c_i$, and $x_i=0$ otherwise. The final conversion derived from impression $i$ is denoted as $v_i$. In the example, $v_i=1$ if the request $i$ brings a deal else $v_i=0$. Thus, the expected total number of conversions is given by $\sum_{i=1}^N x_i v_i$. The advertiser's optimization problem is formulated as follows:
\begin{equation}
\begin{aligned} 
\max_{\mathbf{x}} \quad & \sum_{i=1}^N x_i v_i \\ 
s.t. \quad & \sum_{i=1}^N x_i c_i \le B \\ 
& \frac{\sum_{i=1}^N x_i c_i}{\sum_{i=1}^N x_i v_i} \le tCPA 
\end{aligned}
\end{equation}

To solve this constrained optimization problem, the Lagrange duality method ~\cite{AutoBiddingSurvey,RTB2,USCB} is typically introduced to relax the global constraints. By constructing the dual problem and applying the complementary slackness conditions, the optimal allocation rule can be analyzed locally. Specifically, the condition for winning an impression ($x_i=1$) is determined by comparing the incurred cost $c_i$ against a threshold derived from the value $v_i$, weighted by the dual variables. In the context of a real-time bidding system (assuming a second-price auction mechanism ~\cite{secondprice1, secondprice2}), this threshold theoretically corresponds to the optimal bid $b^*$. Previous studies~\cite{USCB} have derived the optimal bidding strategy for this constrained optimization problem. The closed-form optimal bid is given by:

\begin{equation}
\begin{aligned}
    b^* = \lambda_0 v_i + tCPA  \lambda_1 v_i 
\end{aligned}
\end{equation}

However, it is crucial to note that explicitly calculating this optimal bid requires precise knowledge of the optimal dual parameters $\lambda_1, \lambda_2$ and stable estimation of $v_i$. In real-world online advertising systems, the auction environment and traffic distribution vary rapidly, making static parameter estimation infeasible. Consequently, the auto-bidding problem is typically formulated as a sequential decision-making problem (e.g., a Markov Decision Process) to adaptively learn optimal policies in dynamic environments~\cite{USCB}.

Formally, we define the bidding process as a tuple $({S}, {A}, {P}, {R}, {C})$. At each time step $t$, the bidding agent observes a state $s_t \in {S}$, which typically includes the current ad request features, remaining budget $B_t$, and remaining time. The bidding agent makes an action $a_t \sim \pi_\theta(\cdot|s_t)$, representing the bid price $b_t$. The environment returns a reward $r_t$ (e.g., conversion value) and a cost $c_t$ based on the auction dynamics ${P}$. The optimization objective is to maximize the expected cumulative reward:

\begin{equation}
\begin{aligned}
    \max_{\pi_\theta} \quad & J(\pi_\theta) = \mathbb{E}_{\tau \sim \pi_\theta} \left[ \sum_{t=1}^T  r(s_t, a_t) \right] 
\end{aligned}
\end{equation}

where $\tau = (s_1, a_1, \dots, s_T, a_T)$ denotes the trajectory of the bidding episode. This formulation allows the agent to learn a policy $\pi_\theta$ that dynamically takes an action with maximal long-term reward without relying on static dual parameter assumptions.
Previous work like USCB~\cite{USCB}, MBAB~\cite{MBAB} have demonstrate their capability in auto bidding scenario.

\subsection{Generative Bidding}
Generative decision making reformulates sequential control as conditional sequence modeling rather than explicit value estimation ~\cite{GDM_survey}. A representative approach is the Decision Transformer (DT) ~\cite{DT,Transformer}, which models offline trajectories and predicts actions autoregressively conditioned on historical states, actions, and a target Return-to-Go (RTG):
$$\pi(a_t \mid s_t, \text{RTG}_t, s_{<t}, a_{<t}), \quad \text{RTG}_t=\sum_{k=t}^{T} r_k.$$
This formulation is attractive for auto-bidding because it captures long-term dependencies in logged auction trajectories and avoids unstable value bootstrapping.

Recent generative bidding methods mainly follow two lines. Diffusion-based methods such as AIGB and CBD ~\cite{AIGB,CBD} model state-action trajectories through denoising, while DT-based methods such as GAS and GAVE ~\cite{GAS,GAVE} improve autoregressive bidding with mechanisms such as MCTS and value expansion. We focus on the DT-based line because of its efficient inference and direct fit to sequential bidding logs. However, existing DT-style bidders remain limited by offline imitation: they explore only within the support of logged policies, lack explicit epistemic uncertainty estimation for non-stationary markets ~\cite{uncertainty1,uncertainty2}, and depend heavily on hand-specified RTG targets that are difficult to set on unseen traffic ~\cite{RTGbad,RTGbad2}.

Fine-tuning generative models with reinforcement learning has become an effective way to move beyond imitation, as shown by RLHF/RLVR-style post-training and GRPO ~\cite{InstructGPT,DPO,deepseekR1,minimax,kimi1.5,GRPO}. In auto-bidding, however, direct online exploration is financially risky and auction trajectories are more stochastic than natural language sequences. Thus, a practical generative bidding framework must combine RL-based policy improvement with uncertainty-aware safeguards against unreliable simulated or out-of-distribution trajectories.

\section{Two-stage Uncertainty-aware Generative Auto-Bidding Framework}

\begin{figure*}[h]
  \centering
  \includegraphics[width=\linewidth]{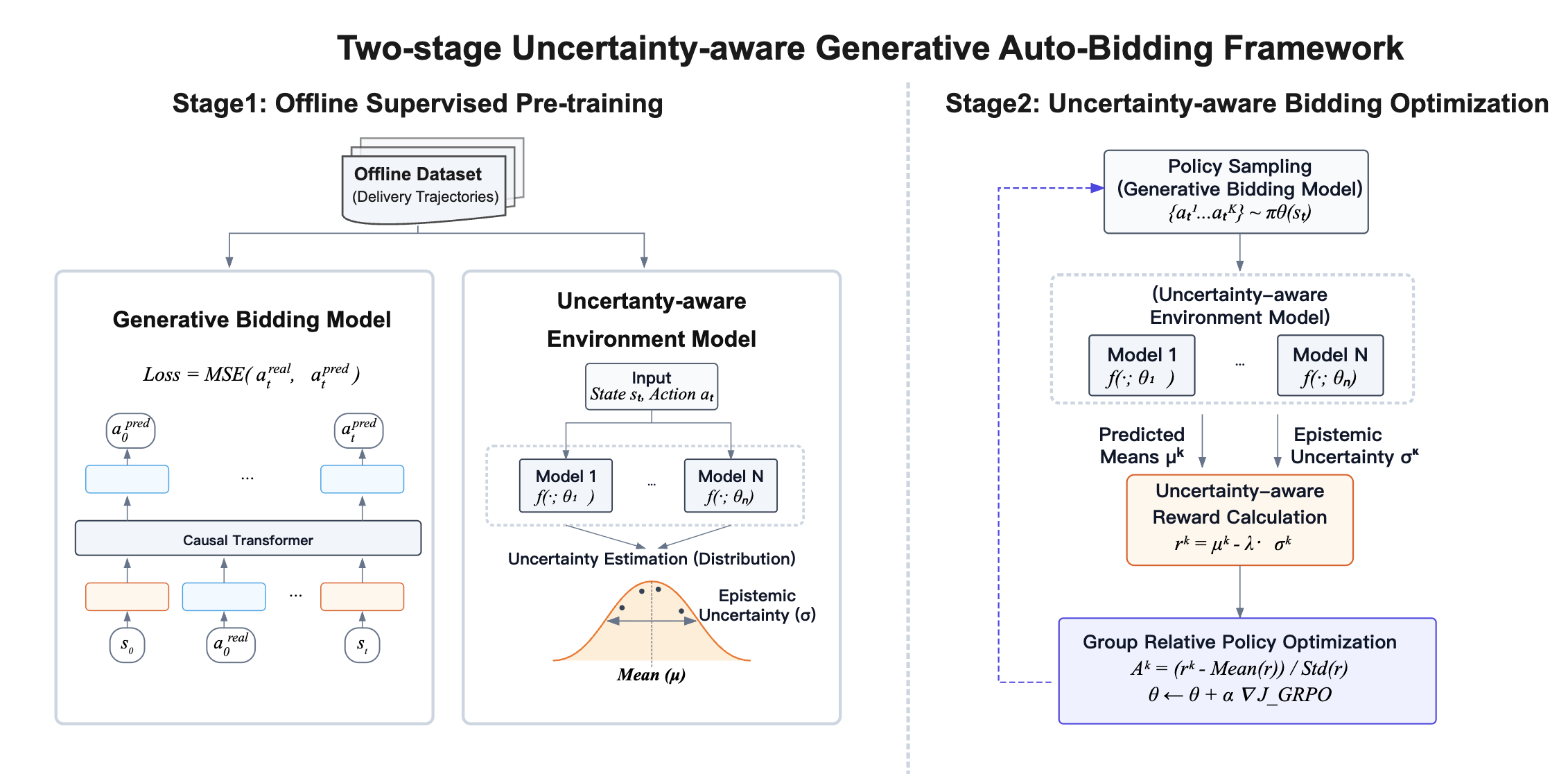}
  \caption{Overview of the TUGB Framework. The pipeline consists of two phases: (1) Offline Supervised Pre-training, where the Generative Bidding Model (GBM) learns a behavioral prior, and the Uncertainty-aware Environment Model (UEM) learns the auction dynamics from the real-world delivery logs; (2) Uncertainty-aware Bidding Optimization, where the agent interacts with the UEM to evolve beyond the offline dataset, guided by uncertainty-penalized rewards.}
  \label{framework}
  \Description{Overview of the TUGB Framework.}
\end{figure*}

To overcome the limitations of static offline datasets and epistemic uncertainty in auto-bidding, we present \textbf{T}wo-stage \textbf{U}ncertainty-aware \textbf{G}enerative \textbf{B}idding framework (TUGB). As shown in Figure~\ref{framework}, TUGB integrates sequence modeling with uncertainty-aware reinforcement learning in two stages. Stage 1 jointly pre-trains a Generative Bidding Model (GBM) as a behavioral prior and an ensemble-based Uncertainty-aware Environment Model (UEM) as a simulator of auction dynamics. Stage 2 fine-tunes the GBM on UEM, where ensemble disagreement penalizes unreliable trajectories and prevents the policy from exploiting out-of-distribution simulation errors. This design is intended to retain the benefits of supervised pre-training while supporting risk-aware policy improvement beyond the logged data.

\subsection{Stage 1: Offline Supervised Pre-training}
In this part, we introduce how the generative bidding model (GBM) and uncertainty environment model (UEM) are pretrained.
Stage 1 learns a behavioral prior and a risk-aware simulator from real-world delivery logs.

\subsubsection{Generative Bidding Model}
Unlike traditional reinforcement learning policies that map isolated states to actions, we treat the bidding process as a sequence modeling problem. We employ a Transformer-based architecture to capture the long-term dependencies within historical bidding trajectories $\tau = (s_0, a_0, \dots, s_T, a_T)$. Standard Decision Transformers (DT) condition the policy on a Return-to-Go (RTG) token to guide generation. However, explicit RTG conditioning in varying auction environments is difficult ~\cite{RTGbad, RTGbad2}, leading to suboptimal performance. Therefore, in TUGB, we alleviate the RTG dependency during the pre-training phase. The Transformer is trained via Behavior Cloning (BC) to minimize the negative log-likelihood of the expert actions given the historical context:
\begin{equation}
    {L}_{BC}(\theta) = - \mathbb{E}_{\tau \sim {D}} \left[ \sum_{t=1}^{T} \log \pi_\theta(a_t | s_{\le t}, a_{<t}) \right]
\end{equation}
where $s_{\le t}$ denotes the sequence of states up to time $t$. This design allows the backbone to focus purely on learning the intrinsic transition dynamics and mimicking the diverse bidding behaviors present in the offline dataset, providing a generalized initialization for the subsequent fine-tuning stage.

\subsubsection{Environment Model}
Enhancing the decision-making capability of the Transformer necessitates a high-fidelity environment model, particularly in real-time bidding where exploring new actions entails significant financial risk. However, accurately modeling the auction landscape is complicated by the inherent epistemic uncertainty of the ad system ~\cite{uncertainty1,uncertainty2}. 
Epistemic uncertainty arises from the non-stationary nature of market dynamics—such as shifting competitor strategies or budget exhaustion, reflecting the model's ignorance regarding Out-of-Distribution (OOD) states. To effectively address this dual challenge, we employ a Probabilistic Ensemble Environment Model to explicitly quantify the epistemic component, thereby enabling risk-aware policy updates that prevent the agent from exploiting simulation errors in unseen market conditions. In addition to uncertainty quantification, employing an ensemble architecture reduces the generalization error of the environment model, leading to more precise state predictions.

We employ an Ensemble model architecture to quantify the epistemic uncertainty. Specifically, we capture  epistemic uncertainty by calculating the divergence among the different sub-models. Based on this formulation, the epistemic uncertainty is calculated as follows:
Let $\{f_{\theta_m}\}_{m=1}^M$ denote the ensemble of $M$ probabilistic models. For a given input $x$, the $m$-th model outputs a Gaussian distribution parameterized by a mean $\mu_{\theta_m}(x)$ and a variance $\sigma_{\theta_m}^2(x)$.

The \textit{epistemic uncertainty}, reflecting the model's lack of knowledge due to environmental dynamics, is quantified as the variance of the means across the ensemble:
\begin{equation}
    \sigma_{epistemic}(x) = \frac{1}{M} \sum_{m=1}^{M} \left( \mu_{\theta_m}(x) - \bar{\mu}(x) \right)^2
\end{equation}
where $\bar{\mu}(x) = \frac{1}{M} \sum_{m=1}^{M} \mu_{\theta_m}(x)$ is the predictive mean of the ensemble.

In auto-bidding scenarios, state variables (such as remaining budget and current CPA) often exhibit high variance and non-linear jumps, making the prediction of absolute states $s_{t+1}$ numerically unstable. To mitigate this, we adopt a residual learning paradigm. The model predicts the state residual $\Delta s_t = s_{t+1} - s_t$ rather than the absolute next state. This approach simplifies the learning landscape by focusing on the local dynamics of change, thereby enhancing the model's sensitivity to action perturbations and improving generalization in continuous spaces. 

\subsection{Stage 2: Uncertainty-aware Bidding Optimization}

Building upon the behavioral prior and environment simulator established in the pre-training stage, we proceed to the fine-tuning phase. The objective is to transcend the limitations of supervised imitation and improve decision-making capabilities of the bidding agent. 

Reinforcement Learning (RL) is widely adopted to refine the reasoning and planning abilities of generative models ~\cite{InstructGPT,deepseekR1,kimi1.5,minimax}. Leveraging our uncertainty-aware environment model, we sample simulated delivery trajectories with varying actions. In the bidding scenario, the optimization objective is explicitly defined; thus, we can evaluate these trajectories using a rule-based reward formulation. For instance, in strict tCPA bidding, the reward $R(\tau)$ for a delivery trajectory $\tau$ could be formalized as ~\cite{AuctionNet}:
\begin{equation}
\begin{aligned}
R(\tau) =  min(\frac{tCPA^\beta}{CPA^\beta},1) * \sum_\tau v 
\end{aligned}
\end{equation}
where $\beta$ is a fixed parameter that take penalty on the tCPA constraint.

Regarding policy optimization, we employ Group Relative Policy Optimization (GRPO)~\cite{GRPO}, a state-of-the-art algorithm in the domain of RL with Verifiable Rewards (RLVR). Unlike standard PPO~\cite{PPO}, GRPO computes advantages by comparing multiple outputs within a group rather than relying on an additional value network. This approach not only alleviates the burden of training a critic, but also enhances the precision of policy updates by using the group mean as a robust baseline.

To mitigate epistemic uncertainty arising from traffic distribution shifts, complex market dynamics, or imperfect modeling of delayed conversions, we employ Uncertainty-aware Environment Model (UEM) to quantify the uncertainty and integrate it as a penalty term directly into the reward signal. This approach alleviates the impact of simulator bias, acting as a critical safety constraint that prevents the policy from chasing unreliable rewards in under-explored regions.  Specifically, the uncertainty-adjusted reward $\hat{R}_i$ and the group-relative advantage $A_i$ for trajectory $\tau_i$ are formulated as:
\begin{equation}
    \hat{R}_i = R(\tau_i) - \lambda \cdot \sigma(\tau_i)
\end{equation}
\begin{equation}
    A_i = \frac{\hat{R}_i - \text{mean}(\{\hat{R}_1, \dots, \hat{R}_G\})}{\text{std}(\{\hat{R}_1, \dots, \hat{R}_G\}) + \epsilon}
\end{equation}
where $\sigma(\tau_i)$ represents the epistemic uncertainty estimated by the ensemble environment model (e.g., variance of predicted returns), and $\lambda$ is the penalty coefficient.
The final objective function is defined as:
\begin{equation}
\begin{aligned}
    J(\theta) = \mathbb{E}_{s \sim D} \bigg[ \frac{1}{G} \sum_{i=1}^G \Big( & \min \big( \rho_i A_i, \text{clip}(\rho_i, 1-\epsilon, 1+\epsilon) A_i \big) \\
    & - \beta_{KL} D_{KL}(\pi_\theta || \pi_{\text{ref}}) \Big) \bigg]
\end{aligned}
\end{equation}
where $\rho_i = \frac{\pi_\theta(\tau_i|s)}{\pi_{\theta_{old}}(\tau_i|s)}$ is the probability ratio, and the KL term prevents the policy from deviating excessively from the reference model $\pi_{\text{ref}}$. 
The pseudocode of total training pipeline of TUGB is shown in Algo~\ref{TUGB_full_alg}.
\begin{algorithm}[t]
\caption{Two-Stage Uncertainty-Aware Generative Auto-Bidding Framework}
\label{TUGB_full_alg}
\begin{algorithmic}[1]
    \STATE \textbf{Stage 1: Offline Supervised Pre-Training}
    \FOR{step $t_1=1$ to $N_{\mathrm{pre}}$}
        \STATE Sample batch $B \sim D$
        \STATE $L_{\mathrm{BC}}(\theta)
            = -\mathbb{E}_{(\tau,a)\sim B}
              [\log \pi_\theta(a\mid\tau)]$
        \STATE $\theta \leftarrow
            \theta-\eta_1\nabla_\theta L_{\mathrm{BC}}$
        \FOR{$m=1$ to $M$}
            \STATE $L_{\mathrm{Model}}(\phi_m)
                = \mathbb{E}_{(\tau,a,r,s')\sim B}
                  [\lVert s'-\hat{s}'_{\phi_m}(\tau,a)\rVert^2]$
            \STATE $\phi_m \leftarrow
                \phi_m-\eta_1\nabla_{\phi_m}L_{\mathrm{Model}}$
        \ENDFOR
    \ENDFOR

    \STATE \textbf{Stage 2: Uncertainty-Aware Bidding Optimization}
    \STATE Initialize $\pi_{\mathrm{ref}}\leftarrow\pi_\theta$
    \FOR{step $t=1$ to $N_{\mathrm{RL}}$}
        \STATE Sample batch of states $S=\{\tau\}\sim D$
        \FOR{each state $\tau\in S$}
            \STATE Sample $G$ actions
                $\{a_i\}_{i=1}^{G}\sim\pi_\theta(\cdot\mid\tau)$
            \STATE Generate trajectories $\tau_i$ using the ensemble
            \FOR{$i=1$ to $G$}
                \STATE Compute $R_i^{\mathrm{mean}}$ and $\sigma_i$
                \STATE $\hat R_i\leftarrow
                    R_i^{\mathrm{mean}}-\lambda\sigma_i$
            \ENDFOR
            \STATE Compute group statistics
                $\mu_{\hat R}$ and $\sigma_{\hat R}$
            \STATE $A_i\leftarrow
                (\hat R_i-\mu_{\hat R})/(\sigma_{\hat R}+\epsilon)$
        \ENDFOR
        \STATE Compute loss $L(\theta)$ via GRPO, Eq.~(9)
        \STATE $\theta\leftarrow
            \theta+\eta\nabla_\theta L(\theta)$
    \ENDFOR
    \RETURN $\pi_\theta^*$
\end{algorithmic}
\end{algorithm}

It is noteworthy that in real-world production deployments, only the Generative Bidding Policy Model is required for inference. The uncertainty-aware environment model, while critical during the fine-tuning stage, does not participate in the online execution flow, thereby maintaining high inference efficiency and minimizing computational overhead.

\section{Experiments}
To validate the effectiveness and deployment viability of the TUGB framework, we conducted extensive evaluations involving both online production A/B tests within JD.com, a leading e-commerce advertising platform and large-scale offline benchmarks. Our experiments are designed to answer the following key research questions: \textbf{RQ1}: Does TUGB deliver substantial improvements in core business metrics  compared to state-of-the-art industrial baselines in real-world online bidding scenarios? \textbf{RQ2}: Can TUGB maintain superior performance and stability across diverse marketing campaigns, fluctuating traffic patterns, and unseen competition dynamics? \textbf{RQ3}: How do the key components specifically the uncertainty aware environment model and the RTG-free mechanism contribute to the model's auto-bidding performance?

\subsection{Online Deployment}

To empirically substantiate the effectiveness of our approach in real-world industrial production settings, we integrated the framework into large-scale real-world bidding system of JD.com, one of the world's largest e-commerce platforms. The evaluation was conducted across two different scenarios:

\begin{itemize}
\item \textbf{JD Search Ad:} The platform's search-ad system. Here, the objective is typically constrained optimization, where advertisers define a target CPA (tCPA) or target ROI (tROI) to maximize conversions.
\item \textbf{JD Food Delivery:} A local services scenario characterized by high temporal dynamics (lunch/dinner peaks) and significantly sparser data compared to retail e-commerce.
\end{itemize}
In both scenarios, the bidding agent must maximize conversion volume while strictly adhering to the advertiser-specified constraints (e.g., $\text{Realized CPA} \le \text{tCPA}$).

\subsubsection{RQ1: Performance on Large-Scale Ad System (JD Search Ad)}

We constructed a large-scale offline dataset derived from online ad delivery logs from JD Search Ad. This dataset comprises over 1 million bidding trajectories and tens of millions of impression opportunities.
We separate each delivery session into $T=24$ time steps. For each step $t$, the state space $s_t$ encompasses multi-dimensional features:
(1) Resource Status: Remaining budget, Remaining  time;
(2) Market Feedback: Realized CPA, Realized CPC, Cost speed, Conversions;
(3) Advertiser Context: tCPA, Product Category, Advertiser type;
The action space $a_t$ is continuous, defined as a scalar bid\_ratio $\alpha \in [0, \alpha_{max}]$.

The training pipeline follows the two-stage paradigm described in Section 3. The Decision Transformer backbone utilizes a context length of $K=24$. We employ the AdamW optimizer with a warm-up strategy.  The ensemble consists of $M=8$ independent Transformer sub-models to capture epistemic uncertainty.
The reward function $R_\tau$ is designed to balance conversion maximization with constraint satisfaction.
\begin{equation}
\begin{aligned}
R_\tau = (1-abs(1-\frac{Real\_CPA}{tCPA}))*conv
\end{aligned}
\end{equation}
As for the parameter choice in Stage 2, the uncertainty penalty coefficient is set as 1 and the GRPO group size is set 16.

We conducted a strictly controlled A/B test against USCB~\cite{USCB}, the current production baseline, over a 7-day observation period. Traffic was randomly split by delivery campaign ID buckets to ensure fairness. As shown in Table~\ref{tab:onlineAB}, TUGB improves the primary business metric Cost by +2.32\% and Target Cost by +3.57\%, while simultaneously improving the constraint satisfaction rate (CSR) by +0.55\%. This simultaneous improvement in both order volume and CSR indicates that TUGB does not simply "spend more" to get conversions; rather, it intelligently reallocates budget from low-efficiency impressions to high-value opportunities, validating the efficacy of the learned value-driven policy.

To further examine the stability of TUGB, we report segment-level results across budget regimes and major product categories. Table~\ref{tab:segment_analysis} shows that TUGB delivers consistent improvements across all budget regimes. The gains are most pronounced in the mid-budget segment, with +3.55\% in Cost and +3.22\% in Target Cost, suggesting that TUGB is particularly effective for advertisers with moderate budgets.
\begin{table}[t]
  \centering
  \caption{Segment-level analysis on JD Search Ad.}
  \label{tab:segment_analysis}
  \setlength{\tabcolsep}{3.5pt}
  \begin{tabular}{llccc}
    \toprule
    Type & Segment & Cost & Target Cost \\
    \midrule
    Budget & 0--1000 & +1.84\% & +4.46\%  \\
    Budget & 1000--10000 & +3.55\% & +3.22\% \\
    Budget & 10000+ & +0.62\% & +0.66\%  \\
    \bottomrule
  \end{tabular}
\end{table}

\subsubsection{RQ2: Robustness and Generalization Capabilities}

A critical requirement for industrial auto-bidding systems is the ability to adapt to non-stationary auction environments and transfer to data-sparse domains. We evaluate TUGB in two representative cases.

\textbf{Case 1: Rapid Adaptation to Mechanism Iterations.}
Since we have deployed TUGB in online system, the JD Search Ad system introduced a bank-account-based pricing mechanism~\cite{mirrokni2016dynamic}, which shifts the bidding objective toward optimizing future conversions. To adapt TUGB to this mechanism update, we simply modified the reward function to $R_\tau = (1- abs(1-\frac{RealCPA_{future}}{tCPA_{future}}))*conv_{future}$ in the fine-tuning stage to align with the new bidding target, while reusing the pre-trained GBM and UEM from Stage 1. Without changing the model architecture or re-collecting a large-scale pre-training dataset, TUGB achieved +2.26\% in Cost and +2.24\% in Target Cost under the new mechanism, as shown in Table~\ref{tab:onlineAB}. This result demonstrates the practical adaptability of TUGB under mechanism changes.

\textbf{Case 2: Cross-Domain Transfer to Data-Sparse Scenarios (JD Food Delivery).} We further evaluated TUGB on JD Food Delivery. In this scenario, we reused the GBM pre-trained on JD Search Ad and trained a new UEM using JD Food Delivery data. This setting is more data-sparse: JD Food Delivery contains only about 100K trajectories, roughly 10 times fewer than the trajectories available in JD Search Ad. Despite the much smaller amount of target-domain data, TUGB achieves a +1.35\% Cost improvement on JD Food Delivery, demonstrating the sample efficiency of the pre-trained GBM prior when adapting to data-sparse domains. Notably, we initially neglected a domain-specific time-sensitivity feature, namely delivery time windows. This caused the transferred model to make suboptimal budget pacing decisions. After incorporating delivery time as a critical feature in the UEM, performance improved substantially. This experience reveals a practical lesson: domain-specific features absent from the source domain must be explicitly identified and incorporated during transfer. 

\begin{table}[t]
  \centering
  \caption{Summary of online deployment result.}
  \label{tab:onlineAB}
  \begin{tabular}{lcc}
    \toprule
    Scenario & Cost & Target Cost \\
    \midrule
    JD Search Ad
    & \makecell{\textbf{+2.32\%}\\$p=0.01$\\ CI=[+0.56,+4.08]}
    & \makecell{\textbf{+3.57\%}\\$p=0.04$\\ CI=[-0.18,+5.96]} \\
    \midrule
    Mechanism Update
    & \makecell{\textbf{+2.26\%}\\$p=0.03$\\ CI=[+0.22,+4.30]}
    & \makecell{\textbf{+2.24\%}\\$p=0.01$\\CI=[+0.54,+3.94]} \\
    \midrule
    JD Food Delivery
    & \makecell{\textbf{+1.35\%}\\$p=0.03$\\ CI=[+0.13,+2.57]}
    & -- \\
    \bottomrule
  \end{tabular}
\end{table}

Overall, Table~\ref{tab:onlineAB} answers RQ1 by confirming significant gains in the main JD Search Ad deployment and answers RQ2 by showing that TUGB remains effective under both a mechanism update and a data-sparse transfer scenario.

\subsection{Offline Evaluation}
The online deployment results above primarily answer RQ1 and RQ2 by measuring business impact and robustness in production. The goal of the offline evaluation is complementary: it provides a reproducible public-benchmark comparison under controlled budget constraints and, more importantly, isolates the contribution of the main TUGB components for RQ3.
\begin{table*}[t]
  \centering
  \caption{Performance comparison against the sota baselines on AuctionNet and AuctionNet-Sparse under varying budget.}
  \label{offline_result}
  \setlength{\tabcolsep}{1.2pt}

  \begin{tabular}{@{}cc*{16}{c}@{}}
    \toprule
    Dataset
    & Budget
    & USCB
    & CQL
    & IQL
    & BCQ
    & DiffBid
    & DT
    & GAS
    & GAVE
    & MOPO
    & MBPO
    & \makecell{TUGB-\\pretrain}
    & \makecell{TUGB-\\PPO}
    & \makecell{TUGB-\\PG}
    & \makecell{TUGB-\\w/o-U}
    & TUGB
    & Improve \\
    \midrule

    \multirow[c]{5}{*}{AuctionNet}
    & 50\%
    & 86 & 113 & 164 & 190 & 54
    & 191 & 193 & 201 & 86 & 67
    & 186 & 191 & 193 & 195
    & \textbf{204} & 1.49\% \\

    & 75\%
    & 135 & 139 & 232 & 259 & 100
    & 265 & 287 & 296 & 128 & 96
    & 259 & 267 & 281 & 274
    & \textbf{302} & 2.03\% \\

    & 100\%
    & 157 & 171 & 281 & 321 & 152
    & 329 & 359 & 376 & 169 & 126
    & 337 & 348 & 367 & 368
    & \textbf{380} & 1.06\% \\

    & 125\%
    & 220 & 201 & 355 & 379 & 193
    & 396 & 409 & 421 & 215 & 156
    & 403 & 411 & 414 & 412
    & \textbf{422} & 0.24\% \\

    & 150\%
    & 281 & 238 & 401 & 429 & 234
    & 450 & 461 & 467 & 256 & 191
    & 467 & 470 & 478 & 474
    & \textbf{483} & 3.43\% \\

    \midrule

    \multirow[c]{5}{*}{\makecell[c]{AuctionNet-\\Sparse}}
    & 50\%
    & 11.5 & 12.8 & 16.5 & 17.7 & 9.9
    & 14.8 & 18.4 & 19.6 & 12.4 & 9.7
    & 16.9 & 18.3 & 19.9 & 18.8
    & \textbf{20.4} & 4.08\% \\

    & 75\%
    & 14.9 & 16.7 & 22.1 & 24.6 & 15.4
    & 22.9 & 27.5 & 28.3 & 16.7 & 14.6
    & 24.7 & 25.3 & 26.1 & 25.4
    & \textbf{28.6} & 1.06\% \\

    & 100\%
    & 17.5 & 22.2 & 30.0 & 31.1 & 19.5
    & 29.6 & 36.1 & 37.2 & 25.6 & 15.8
    & 31.0 & 32.6 & 36.1 & 34.3
    & \textbf{37.4} & 0.54\% \\

    & 125\%
    & 26.7 & 28.6 & 37.1 & 34.2 & 25.3
    & 34.3 & 40.0 & 42.7 & 29.3 & 24.6
    & 38.4 & 40.6 & 42.8 & 42.3
    & \textbf{44.0} & 3.04\% \\

    & 150\%
    & 31.3 & 35.8 & 43.1 & 37.9 & 30.8
    & 44.5 & 46.5 & 47.4 & 36.4 & 31.6
    & 42.5 & 42.3 & 47.6 & 44.7
    & \textbf{48.2} & 1.69\% \\

    \bottomrule
  \end{tabular}
\end{table*}

\subsubsection{Evaluation Setup}
To ensure reproducibility and facilitate fair comparison, we utilize AuctionNet~\cite{AuctionNet}, a large-scale public benchmark released by Alibaba that faithfully simulates the complexity of real-world ad auctions. The environment simulates a highly competitive market with 48 advertisers competing for impression opportunities across 21 delivery periods. Each period consists of 48 time steps involving approximately 500,000 impression requests. Our study utilizes two specific subsets: (1) \textit{AuctionNet-Standard}, which contains comprehensive bidding trajectories representing typical market conditions; and (2) \textit{AuctionNet-Sparse}, a challenging variant characterized by significantly reduced conversion rates (CVR), designed to test model robustness in data-sparse scenarios.

We benchmark TUGB against representative baselines from four categories. 
First, we include offline RL and production bidding methods, including IQL~\cite{IQL}, CQL~\cite{CQL}, BCQ~\cite{BCQ}, and USCB~\cite{USCB}, where USCB is a constrained bidding algorithm deployed in our production system. 
Second, we compare with model-based RL methods, including MBPO~\cite{MBPO} and MOPO~\cite{MOPO}, to evaluate whether TUGB brings benefits beyond standard model-based policy optimization and uncertainty-penalized rollouts. 
Third, we include recent generative bidding methods, including DiffBid~\cite{AIGB}, vanilla Decision Transformer~\cite{DT}, GAS~\cite{GAS}, and GAVE~\cite{GAVE}. 
Finally, we conduct ablation studies with TUGB-pretrain, TUGB-PPO, TUGB-PG, and TUGB-w/o-U to isolate the effects of supervised generative pre-training, GRPO-style optimization, vanilla policy-gradient optimization, and uncertainty-aware ensemble modeling.

We adopt score ($R$) defined in AuctionNet~\cite{AuctionNet} as the decisive metric to evaluate bidding performance. All experiments are conducted on NVIDIA P40 GPUs using a fixed batch size of 200. The model parameters are optimized through AdamW with a learning rate of $1\times 10^{-5}$. To ensure a fair comparison, all Transformer-based variants (DT, GAVE, and TUGB) share a unified backbone architecture consisting of 3 transformer layers with 16 attention heads.

\subsubsection{RQ3: Component Contribution Analysis}
Table~\ref{offline_result} answers RQ3 by comparing the full TUGB model with both external baselines and internal variants. TUGB achieves the best reward under every budget setting on both AuctionNet and AuctionNet-Sparse, improving over the strongest non-TUGB baseline by up to 3.43\% and 4.08\%, respectively. The ablation results further identify the contribution of each core component.

First, the uncertainty-aware environment model is essential for stable policy improvement. TUGB-w/o-U consistently underperforms the full model, especially on AuctionNet-Sparse, indicating that a single environment model is more vulnerable to value overestimation under sparse feedback. The ensemble UEM mitigates this issue by penalizing uncertain rollouts and enabling safer exploration. This conclusion is also supported by an online ablation study in JD Search Ad. Replacing the ensemble UEM with a single environment model yields +1.32\% Cost and +2.09\% Target Cost, compared with +2.32\% and +3.57\% from the full TUGB model. The gap corresponds to approximately +1.00\% in Cost and +1.48\% in Target Cost, confirming that the ensemble-based uncertainty mechanism contributes additional production gains by reducing value overestimation and improving policy optimization robustness.

Second, the RTG-free design provides a stronger starting point than directly applying a standard Decision Transformer. TUGB-pretrain outperforms DT in most budget settings, suggesting that fixed RTG conditioning is unreliable in stochastic auto-bidding. However, TUGB further improves over TUGB-pretrain, showing that RTG-free pretraining and uncertainty-aware policy optimization are complementary.

We further examine whether the above conclusions are sensitive to key hyper-parameters. As shown in Table~\ref{tab:sensitivity}, TUGB remains stable around the deployment configuration: a moderate uncertainty penalty performs best, while larger $M$ and $G$ improve reward with diminishing returns and higher rollout cost. We therefore use $\lambda=1$, $M=8$, and $G=16$ in deployment.

\begin{table}[t]
  \centering
  \caption{Sensitivity on AuctionNet with 100\% budget. Entries denote
  parameter value/$R$; $\dagger$ marks the default.}
  \label{tab:sensitivity}
  \begin{tabular}{@{}ccccc@{}}
    \toprule
    Param. & \multicolumn{4}{c}{Value / $R$} \\
    \midrule
    $\lambda$
      & 0/368 & 0.5/376 & $\mathbf{1}^{\dagger}$/\textbf{380} & 2/344 \\
    $M$
      & 2/356 & 4/366 & $8^{\dagger}$/380 & \textbf{16}/\textbf{384} \\
    $G$
      & 2/364 & 4/368 & 8/373 & $\mathbf{16}^{\dagger}$/\textbf{380} \\
    \bottomrule
  \end{tabular}
\end{table}

\section{Conclusion and Future Work}
We presented TUGB, which combines RTG-free generative pre-training with uncertainty-aware policy optimization. Production deployment and AuctionNet experiments demonstrate improved performance, robustness, and cross-domain transfer. Future work will extend TUGB to multi-objective bidding and more adaptive market dynamics.

\section{GenAI Usage Disclosure}
The authors used generative AI tools to assist with language polishing, conciseness editing, and formatting suggestions. All technical ideas, experimental design, results, analysis, and conclusions were created, verified, and approved by the authors.

\clearpage
\bibliographystyle{ACM-Reference-Format}
\balance
\bibliography{citiations}

@inproceedings{AIGB,
  title={Generative auto-bidding via conditional diffusion modeling},
  author={Guo, Jiayan and Huo, Yusen and Zhang, Zhilin and Wang, Tianyu and Yu, Chuan and Xu, Jian and Zheng, Bo and Zhang, Yan},
  booktitle={Proceedings of ACM SIGKDD Conference on Knowledge Discovery and Data Mining},
  pages={5038--5049},
  year={2024}
}

@article{DT,
  title={Decision transformer: Reinforcement learning via sequence modeling},
  author={Chen, Lili and Lu, Kevin and Rajeswaran, Aravind and Lee, Kimin and Grover, Aditya and Laskin, Misha and Abbeel, Pieter and Srinivas, Aravind and Mordatch, Igor},
  journal={Proceedings of Advances in neural information processing systems},
  volume={34},
  pages={15084--15097},
  year={2021}
}

@article{Transformer,
  title={Attention is all you need},
  author={Vaswani, Ashish and Shazeer, Noam and Parmar, Niki and Uszkoreit, Jakob and Jones, Llion and Gomez, Aidan N and Kaiser, {\L}ukasz and Polosukhin, Illia},
  journal={Proceedings of Advances in neural information processing systems},
  volume={30},
  year={2017}
}

@article{GDM_survey,
  title={Generative models in decision making: A survey},
  author={Li, Yinchuan and Shao, Xinyu and Zhang, Jianping and Wang, Haozhi and Brunswic, Leo Maxime and Zhou, Kaiwen and Dong, Jiqian and Guo, Kaiyang and Li, Xiu and Chen, Zhitang and others},
  journal={arXiv preprint arXiv:2502.17100},
  year={2025}
}

@inproceedings{USCB,
author = {He, Yue and Chen, Xiujun and Wu, Di and Pan, Junwei and Tan, Qing and Yu, Chuan and Xu, Jian and Zhu, Xiaoqiang},
title = {A Unified Solution to Constrained Bidding in Online Display Advertising},
year = {2021},
booktitle = {Proceedings of ACM SIGKDD Conference on Knowledge Discovery \& Data Mining},
pages = {2993–3001},
}

@inproceedings{GAVE,
  title={Generative auto-bidding with value-guided explorations},
  author={Gao, Jingtong and Li, Yewen and Mao, Shuai and Jiang, Peng and Jiang, Nan and Wang, Yejing and Cai, Qingpeng and Pan, Fei and Jiang, Peng and Gai, Kun and others},
  booktitle={Proceedings of International ACM SIGIR Conference on Research and Development in Information Retrieval},
  pages={244--254},
  year={2025}
}

@inproceedings{GAS,
  title={GAS: Generative Auto-bidding with Post-training Search},
  author={Li, Yewen and Mao, Shuai and Gao, Jingtong and Jiang, Nan and Xu, Yunjian and Cai, Qingpeng and Pan, Fei and Jiang, Peng and An, Bo},
  booktitle={Companion Proceedings of ACM on Web Conference},
  pages={315--324},
  year={2025}
}

@inproceedings{likaiyuanEBaReT,
  title={EBaReT: Expert-guided Bag Reward Transformer for Auto Bidding},
  author={Li, Kaiyuan and Wang, Pengyu and Peng, Yunshan and Yuan, Pengjia and Zeng, Yanxiang and Xiang, Rui and Cheng, Yanhua and Liu, Xialong and Jiang, Peng},
  booktitle={Companion Proceedings of ACM on Web Conference},
  pages={1104--1108},
  year={2025}
}

@article{CQL,
  title={Conservative q-learning for offline reinforcement learning},
  author={Kumar, Aviral and Zhou, Aurick and Tucker, George and Levine, Sergey},
  journal={Proceedings of Advances in neural information processing systems},
  volume={33},
  pages={1179--1191},
  year={2020}
}

@inproceedings{IQL,
  title={Offline Reinforcement Learning with Implicit Q-Learning},
  author={Ilya Kostrikov and Ashvin Nair and Sergey Levine},
  booktitle={Proceedings of International Conference on Learning Representations},
  year={2022}
}

@inproceedings{BCQ,
  title={Off-policy deep reinforcement learning without exploration},
  author={Fujimoto, Scott and Meger, David and Precup, Doina},
  booktitle={Proceedings of International conference on machine learning},
  pages={2052--2062},
  year={2019},
  organization={PMLR}
}

@article{CBD,
  title={Generative Auto-Bidding in Large-Scale Competitive Auctions via Diffusion Completer-Aligner},
  author={Li, Yewen and Gao, Jingtong and Jiang, Nan and Mao, Shuai and An, Ruyi and Pan, Fei and Zhao, Xiangyu and An, Bo and Cai, Qingpeng and Jiang, Peng},
  journal={arXiv preprint arXiv:2509.03348},
  year={2025}
}

@article{PPO,
  title={Proximal policy optimization algorithms},
  author={Schulman, John and Wolski, Filip and Dhariwal, Prafulla and Radford, Alec and Klimov, Oleg},
  journal={arXiv preprint arXiv:1707.06347},
  year={2017}
}

@article{onlineAD,
  title={The online advertising industry: Economics, evolution, and privacy},
  author={Evans, David S},
  journal={Journal of economic perspectives},
  volume={23},
  number={3},
  pages={37--60},
  year={2009},
  publisher={American Economic Association}
}

@inproceedings{RTB2,
  title={Real-time bidding: A new frontier of computational advertising research},
  author={Wang, Jun and Yuan, Shuai},
  booktitle={Proceedings of ACM International Conference on Web Search and Data Mining},
  pages={415--416},
  year={2015}
}

@inproceedings{RTB1,
  title={Real-time bidding algorithms for performance-based display ad allocation},
  author={Chen, Ye and Berkhin, Pavel and Anderson, Bo and Devanur, Nikhil R},
  booktitle={Proceedings of ACM SIGKDD international conference on Knowledge discovery and data mining},
  pages={1307--1315},
  year={2011}
}

@article{RTGbad,
  title={When does return-conditioned supervised learning work for offline reinforcement learning?},
  author={Brandfonbrener, David and Bietti, Alberto and Buckman, Jacob and Laroche, Romain and Bruna, Joan},
  journal={Proceedings of Advances in Neural Information Processing Systems},
  volume={35},
  pages={1542--1553},
  year={2022}
}

@article{RTGbad2,
  title={You can’t count on luck: Why decision transformers and rvs fail in stochastic environments},
  author={Paster, Keiran and McIlraith, Sheila and Ba, Jimmy},
  journal={Proceedings of Advances in neural information processing systems},
  volume={35},
  pages={38966--38979},
  year={2022}
}

@article{AutoBiddingSurvey,
  title={Auto-bidding and auctions in online advertising: A survey},
  author={Aggarwal, Gagan and Badanidiyuru, Ashwinkumar and Balseiro, Santiago R and Bhawalkar, Kshipra and Deng, Yuan and Feng, Zhe and Goel, Gagan and Liaw, Christopher and Lu, Haihao and Mahdian, Mohammad and others},
  journal={ACM SIGecom Exchanges},
  volume={22},
  number={1},
  pages={159--183},
  year={2024},
  publisher={ACM New York, NY, USA}
}

@article{OODdanger,
  title={Why so pessimistic? estimating uncertainties for offline rl through ensembles, and why their independence matters},
  author={Ghasemipour, Kamyar and Gu, Shixiang Shane and Nachum, Ofir},
  journal={Proceedings of Advances in Neural Information Processing Systems},
  volume={35},
  pages={18267--18281},
  year={2022}
}

@inproceedings{xiyaopolicygap,
  title={Live in the moment: Learning dynamics model adapted to evolving policy},
  author={Wang, Xiyao and Wongkamjan, Wichayaporn and Jia, Ruonan and Huang, Furong},
  booktitle={Proceedings of International Conference on Machine Learning},
  pages={36470--36493},
  year={2023},
  organization={PMLR}
}

@article{offlineRLsurvey,
  title={A survey on offline reinforcement learning: Taxonomy, review, and open problems},
  author={Prudencio, Rafael Figueiredo and Maximo, Marcos ROA and Colombini, Esther Luna},
  journal={IEEE Transactions on Neural Networks and Learning Systems},
  volume={35},
  number={8},
  pages={10237--10257},
  year={2023},
  publisher={IEEE}
}

@inproceedings{GapinofflineRL,
  title={The Generalization Gap in Offline Reinforcement Learning},
  author={Ishita Mediratta and Qingfei You and Minqi Jiang and Roberta Raileanu},
  booktitle={Proceedings of International Conference on Learning Representations},
  year={2024}
}

@inproceedings{secondprice1,
  title={General auction mechanism for search advertising},
  author={Aggarwal, Gagan and Muthukrishnan, Shan and P{\'a}l, D{\'a}vid and P{\'a}l, Martin},
  booktitle={Proceedings of international conference on World wide web},
  pages={241--250},
  year={2009}
}

@article{secondprice2,
  title={Internet advertising and the generalized second-price auction: Selling billions of dollars worth of keywords},
  author={Edelman, Benjamin and Ostrovsky, Michael and Schwarz, Michael},
  journal={American economic review},
  volume={97},
  number={1},
  pages={242--259},
  year={2007},
  publisher={American Economic Association}
}

@article{InstructGPT,
  title={Training language models to follow instructions with human feedback},
  author={Ouyang, Long and Wu, Jeffrey and Jiang, Xu and Almeida, Diogo and Wainwright, Carroll and Mishkin, Pamela and Zhang, Chong and Agarwal, Sandhini and Slama, Katarina and Ray, Alex and others},
  journal={Proceedings of Advances in neural information processing systems},
  volume={35},
  pages={27730--27744},
  year={2022}
}

@article{DPO,
  title={Direct preference optimization: Your language model is secretly a reward model},
  author={Rafailov, Rafael and Sharma, Archit and Mitchell, Eric and Manning, Christopher D and Ermon, Stefano and Finn, Chelsea},
  journal={Proceedings of Advances in neural information processing systems},
  volume={36},
  pages={53728--53741},
  year={2023}
}

@article{GRPO,
  title={Deepseekmath: Pushing the limits of mathematical reasoning in open language models},
  author={Shao, Zhihong and Wang, Peiyi and Zhu, Qihao and Xu, Runxin and Song, Junxiao and Bi, Xiao and Zhang, Haowei and Zhang, Mingchuan and Li, YK and Wu, Yang and others},
  journal={arXiv preprint arXiv:2402.03300},
  year={2024}
}

@article{deepseekR1,
  title={DeepSeek-R1 incentivizes reasoning in LLMs through reinforcement learning},
  author={Guo, Daya and Yang, Dejian and Zhang, Haowei and Song, Junxiao and Wang, Peiyi and Zhu, Qihao and Xu, Runxin and Zhang, Ruoyu and Ma, Shirong and Bi, Xiao and others},
  journal={Nature},
  volume={645},
  number={8081},
  pages={633--638},
  year={2025},
  publisher={Nature Publishing Group UK London}
}

@inproceedings{MBAB,
  title={Model-based reinforcement learning for auto-bidding in display advertising},
  author={Chen, Shuang and Xu, Qisen and Zhang, Liang and Jin, Yongbo and Li, Wenhao and Mo, Linjian},
  booktitle={Proceedings of International Conference on Autonomous Agents and Multiagent Systems},
  pages={1560--1568},
  year={2023}
}

@article{minimax,
  title={MiniMax-M1: Scaling Test-Time Compute Efficiently with Lightning Attention},
  author={Chen, Aili and Li, Aonian and Gong, Bangwei and Jiang, Binyang and Fei, Bo and Yang, Bo and Shan, Boji and Yu, Changqing and Wang, Chao and Zhu, Cheng and others},
  journal={arXiv preprint arXiv:2506.13585},
  year={2025}
}

@article{AuctionNet,
  title={Auctionnet: A novel benchmark for decision-making in large-scale games},
  author={Su, Kefan and Huo, Yusen and Zhang, Zhilin and Dou, Shuai and Yu, Chuan and Xu, Jian and Lu, Zongqing and Zheng, Bo},
  journal={Proceedings of Advances in Neural Information Processing Systems},
  volume={37},
  pages={94428--94452},
  year={2024}
}

@article{kimi1.5,
  title={Kimi k1. 5: Scaling reinforcement learning with llms},
  author={Team, Kimi and Du, Angang and Gao, Bofei and Xing, Bowei and Jiang, Changjiu and Chen, Cheng and Li, Cheng and Xiao, Chenjun and Du, Chenzhuang and Liao, Chonghua and others},
  journal={arXiv preprint arXiv:2501.12599},
  year={2025}
}

@article{MBPO,
  title={When to trust your model: Model-based policy optimization},
  author={Janner, Michael and Fu, Justin and Zhang, Marvin and Levine, Sergey},
  journal={Proceedings of Advances in neural information processing systems},
  volume={32},
  year={2019}
}

@article{MOPO,
  title={Mopo: Model-based offline policy optimization},
  author={Yu, Tianhe and Thomas, Garrett and Yu, Lantao and Ermon, Stefano and Zou, James Y and Levine, Sergey and Finn, Chelsea and Ma, Tengyu},
  journal={Proceedings of Advances in Neural Information Processing Systems},
  volume={33},
  pages={14129--14142},
  year={2020}
}

@article{uncertainty1,
  title={Simple and scalable predictive uncertainty estimation using deep ensembles},
  author={Lakshminarayanan, Balaji and Pritzel, Alexander and Blundell, Charles},
  journal={Proceedings of Advances in neural information processing systems},
  volume={30},
  year={2017}
}

@inproceedings{uncertainty2,
  title={Decomposition of uncertainty in Bayesian deep learning for efficient and risk-sensitive learning},
  author={Depeweg, Stefan and Hernandez-Lobato, Jose-Miguel and Doshi-Velez, Finale and Udluft, Steffen},
  booktitle={Proceedings of International conference on machine learning},
  pages={1184--1193},
  year={2018},
  organization={PMLR}
}

@inproceedings{mirrokni2016dynamic,
  title={Dynamic Auctions with Bank Accounts},
  author={Mirrokni, Vahab S. and Paes Leme, Renato and Tang, Pingzhong and Zuo, Song},
  booktitle={Proceedings of the Twenty-Fifth International Joint Conference on Artificial Intelligence},
  pages={387--393},
  year={2016}
}

\end{document}